\documentclass{article}
\usepackage[numbers]{natbib}

\usepackage{arxiv}

\usepackage[utf8]{inputenc} 
\usepackage[T1]{fontenc}    
\usepackage{hyperref}       
\usepackage{url}            
\usepackage{booktabs}       
\usepackage{amsfonts}       
\usepackage{amsmath}        
\usepackage{nicefrac}       
\usepackage{microtype}      
\usepackage{lipsum}         
\usepackage{graphicx}
\usepackage{natbib}
\usepackage{doi}

\title{
Fine-Tuning Without Forgetting: Adaptation of YOLOv8 Preserves COCO Performance
}

\author{ {\hspace{1mm}Vishal Gandhi} \\
	Joyspace AI\\
	\texttt{vishal@joyspace.ai} \\
	\And
	{\hspace{1mm}Sagar Gandhi} \\
	Joyspace AI\\
	\texttt{sagar@joyspace.ai} \\
}

\renewcommand{\headeright}{} 
\renewcommand{\undertitle}{}
\renewcommand{\shorttitle}{}

\begin{document}
\maketitle
\pagestyle{plain} 

\begin{abstract}
The success of large pre-trained object detectors hinges on their adaptability to diverse downstream tasks. While fine-tuning is the standard adaptation method, specializing these models for challenging fine-grained domains necessitates careful consideration of feature granularity. The critical question remains: how deeply should the pre-trained backbone be fine-tuned to optimize for the specialized task without incurring catastrophic forgetting of the original general capabilities? Addressing this, we present a systematic empirical study evaluating the impact of fine-tuning depth. We adapt a standard YOLOv8n model to a custom, fine-grained fruit detection dataset by progressively unfreezing backbone layers (freeze points at layers 22, 15, and 10) and training. Performance was rigorously evaluated on both the target fruit dataset and, using a dual-head evaluation architecture, on the original COCO validation set. Our results demonstrate unequivocally that deeper fine-tuning (unfreezing down to layer 10) yields substantial performance gains (e.g., +10\% absolute mAP50) on the fine-grained fruit task compared to only training the head. Strikingly, this significant adaptation and specialization resulted in negligible performance degradation (<0.1\% absolute mAP difference) on the COCO benchmark across all tested freeze levels. We conclude that adapting mid-to-late backbone features is highly effective for fine-grained specialization. Critically, our results demonstrate this adaptation can be achieved without the commonly expected penalty of catastrophic forgetting, presenting a compelling case for exploring deeper fine-tuning strategies, particularly when targeting complex domains or when maximizing specialized performance is paramount.
\end{abstract}


\section{Introduction}

Object detection stands as a cornerstone of modern computer vision, enabling machines to locate and identify objects within images and videos. The advent of deep learning, particularly Convolutional Neural Networks (CNNs), has propelled remarkable progress, with architectures like the YOLO (You Only Look Once) family~\cite{Ref-YOLOv1, Ref-YOLOv3} achieving state-of-the-art performance across diverse benchmarks. These models, typically pre-trained on large-scale, general-purpose datasets like COCO~\cite{Ref-COCO}, serve as powerful foundational tools, capturing a rich hierarchy of visual features applicable to a wide array of tasks.

However, many real-world applications demand specialization beyond the scope of general datasets. Adapting these pre-trained models to specific, often fine-grained domains,such as identifying particular plant species, detecting subtle manufacturing defects, recognizing specific retail products, or analyzing nuanced medical imagery,presents a significant challenge. In such domains, the visual distinctions between classes can be minimal, requiring models to leverage highly discriminative, often subtle, features that may not be emphasized during general pre-training.

The standard approach to bridge this gap is transfer learning, specifically through fine-tuning. This involves adapting the weights of the pre-trained model using data from the target domain. Yet, this raises critical questions regarding the optimal strategy: Should one simply fine-tune the final classification and regression layers (the \textit{head}), preserving the vast majority of the pre-trained feature extractor (the \textit{backbone})? This is computationally efficient and often considered safer against overfitting to potentially smaller target datasets. Or should one fine-tune deeper into the backbone, allowing the model to adapt its feature representations more thoroughly to the nuances of the new domain? While potentially yielding better specialized performance, deeper fine-tuning is computationally more expensive and notoriously carries the risk of catastrophic forgetting~\cite{Ref-FrenchForgetting, Ref-EWC} i.e. the degradation of the model's performance on the original pre-training task as it adapts to the new one. Determining the appropriate depth of fine-tuning, balancing adaptation against potential forgetting and computational cost, remains a critical practical consideration.

While the trade-offs of fine-tuning depth are conceptually understood, there is a relative scarcity of systematic empirical studies directly investigating this spectrum for fine-grained object detection, particularly those that simultaneously quantify the performance impact on both the specialized target task and the original general-purpose source task. Existing guidance often relies on heuristics, and simplistic applications (e.g., merely appending a new head without considering backbone adaptation) may fail to unlock the full potential of transfer learning for complex, fine-grained problems. This gap motivates our work: a focused empirical investigation into the relationship between fine-tuning depth, specialized performance gains, and catastrophic forgetting in a practical fine-grained detection scenario.

In this paper, we make the following contributions:
\begin{itemize}
    \item We present a systematic empirical study adapting a standard pre-trained YOLOv8n model to a custom, fine-grained fruit detection dataset by varying the depth of backbone fine-tuning (layers frozen up to index 22, 15, and 10).
    \item We rigorously evaluate the performance impact on both the target fine-grained fruit detection task and, using a dual-head architecture for consistent evaluation, on the original COCO general object detection task.
    \item We quantify the significant performance improvements achieved on the target task through deeper fine-tuning.
    \item We quantify and analyze the catastrophic forgetting aspect, finding negligible performance degradation on the source COCO task across all fine-tuning depths explored in our setup.
    \item We provide empirical evidence analyzing the trade-offs associated with adapting features at different levels of granularity for fine-grained object detection.
\end{itemize}

The remainder of this paper is structured as follows: Section 2 discusses related work in transfer learning, fine-grained recognition, and catastrophic forgetting. Section 3 details our methodology, including the base model, dataset, fine-tuning protocol, and evaluation strategy. Section 4 presents our experimental setup and quantitative results. 

\section{Related Work}
This work intersects with several established areas of computer vision research, primarily transfer learning for object detection, fine-grained visual categorization, and catastrophic forgetting in neural networks.

\subsection{Transfer Learning in Object Detection}
The paradigm of transfer learning, particularly leveraging models pre-trained on large-scale datasets like ImageNet~\cite{Ref-ImageNet} or COCO~\cite{Ref-COCO}, has become foundational to modern object detection~\cite{Ref-TransferSurvey1}. Pre-training captures robust hierarchical features that often generalize well, significantly reducing the need for vast amounts of labeled data and computational resources for downstream tasks~\cite{Ref-RCNN, Ref-TransferBenefit}. Common adaptation strategies involve either using the pre-trained model as a fixed feature extractor and training only a new task-specific head, or fine-tuning some or all layers of the pre-trained network on the target dataset~\cite{Ref-YosinskiFT}. Fully fine-tuning all layers can yield high performance but risks overfitting on smaller datasets and losing valuable general features. Conversely, freezing the entire backbone is computationally cheaper but may limit adaptability, especially if the target domain differs significantly from the source~\cite{Ref-TransferSurvey2}. Our work investigates the middle ground, systematically exploring the impact of partial backbone fine-tuning depth on both target task performance and source task retention, a practical trade-off space often governed by heuristics rather than detailed empirical analysis in specific application contexts.

\subsection{Fine-Grained Visual Categorization (FGVC)}
Fine-Grained Visual Categorization presents unique challenges due to subtle inter-class variations and potentially large intra-class variations~\cite{Ref-FGVC-Challenge1, Ref-FGVC-Survey}. Distinguishing between species of birds, models of cars, or types of fruits, as in our case, often requires capturing highly localized and discriminative features that may be overlooked by models trained for broader categorization~\cite{Ref-FGVC-Challenge2}. Specialized FGVC methods have been proposed, including those employing attention mechanisms to focus on relevant parts~\cite{Ref-AttentionFGVC}, higher-order feature interactions like Bilinear CNNs~\cite{Ref-BilinearCNN}, or specialized part-localization modules~\cite{Ref-PartBasedFGVC}. While these methods advance the state-of-the-art in FGVC benchmarks, they often introduce architectural complexity or specific training procedures. In contrast, our work does not propose a new FGVC architecture. Instead, we analyze the capability of a standard, widely-used object detector (YOLOv8n) architecture to handle a fine-grained task simply through controlled fine-tuning of its existing layers, providing insights into how adaptable standard feature hierarchies are for fine-grained distinctions.

\subsection{Catastrophic Forgetting and Continual Learning}
When a pre-trained neural network is fine-tuned on a new task, it often suffers from catastrophic forgetting – a drastic decrease in performance on the original task(s) it was trained on~\cite{Ref-McCloskeyForgetting, Ref-FrenchForgetting}. This phenomenon poses a major obstacle for continual learning, where models are expected to learn new tasks sequentially without losing prior knowledge. Various strategies have been developed to mitigate forgetting, broadly categorized as: (1) Regularization-based methods, which add constraints to the loss function to penalize changes to weights deemed important for previous tasks (e.g., Elastic Weight Consolidation (EWC)~\cite{Ref-EWC}, Synaptic Intelligence~\cite{Ref-SI}); (2) Rehearsal-based methods, which store and replay samples from previous tasks during training on new tasks~\cite{Ref-Rehearsal1, Ref-GEM}; and (3) Architecture-based methods, which dynamically allocate model parameters to different tasks~\cite{Ref-ArchitectureCL}. Our work provides an interesting contrast: we observe minimal catastrophic forgetting on the source COCO task despite significant weight updates during fine-tuning on the new fruit task, without employing any explicit forgetting mitigation techniques. This observation motivates further analysis into the inherent resilience of certain architectures or training setups and the conditions under which standard fine-tuning might suffice without significant knowledge loss.

\subsection{YOLO Object Detection Models}
The YOLO (You Only Look Once) family represents a lineage of highly influential real-time object detectors, starting with the original proposal by Redmon et al.~\cite{Ref-YOLOv1}. Subsequent versions, such as YOLOv3~\cite{Ref-YOLOv3} and the numerous iterations often colloquially referred to as YOLOv4/v5~\cite{Ref-YOLOv4, Ref-YOLOv5}, introduced architectural improvements (e.g., feature pyramids, anchor boxes, improved backbones) enhancing accuracy and flexibility. YOLOv8~\cite{Ref-YOLOv8Ultralytics} continues this evolution, incorporating recent advancements like a CSPDarknet backbone, anchor-free detection heads, and optimized training strategies. We selected YOLOv8n, the smallest variant, as our base model due to its widespread adoption, strong baseline performance, and computational efficiency, making it a relevant choice for studying practical transfer learning scenarios.

\section{Methodology}

This section details the base model architecture, the target dataset used for fine-tuning, the experimental fine-tuning protocols employed, and the evaluation strategies used to assess performance on both the target and source tasks.

\subsection{Base Model}

We utilize YOLOv8n~\cite{Ref-YOLOv1, Ref-YOLOv8Paper, Ref-YOLOv8Ultralytics} as our base object detection model. YOLOv8n represents a state-of-the-art iteration in the YOLO family, featuring a CSPDarknet-based backbone for efficient feature extraction, a PANet-inspired neck for feature aggregation across scales~\cite{Ref-BGSYOLO}, and a decoupled, anchor-free detection head. The specific model used was pre-trained on the large-scale COCO (Common Objects in Context) dataset~\cite{Ref-COCO}, equipping it with robust general-purpose visual representations relevant to 80 common object categories.

\subsection{Target Dataset: Fine-Grained Fruits}

To simulate a realistic application scenario requiring domain specialization, we adapt the base model to a fine-grained fruit detection task. Many real-world applications, such as automated fridge content analysis or retail checkout systems, require recognizing a broader or more specific set of object classes than those covered by general datasets like COCO. While COCO includes some fruits (e.g., apple, banana, orange), extending this capability is often necessary.

For our experiments, we created a specialized dataset by selecting and filtering a subset of classes from the publicly available \textit{whatsInYourFridge} dataset~\cite{Ref-FridgeDataset}. This dataset provides images relevant to typical household contents. We specifically curated a 6-class fruit dataset consisting of \textit{orange}, \textit{pear}, \textit{pineapple}, \textit{plum}, \textit{strawberries}, and \textit{watermelon}. This selection presents a fine-grained challenge, requiring the model to distinguish between visually similar fruit types.

The original dataset annotations were preprocessed: only labels corresponding to these six classes were retained. The class indices were remapped sequentially from 0 to 5. Any images containing none of the target fruit classes after filtering were excluded from the dataset. The original dataset's train, validation, and test splits were maintained for our subset.

\begin{figure}[h]
    \centering
    \includegraphics[width=0.45\textwidth]{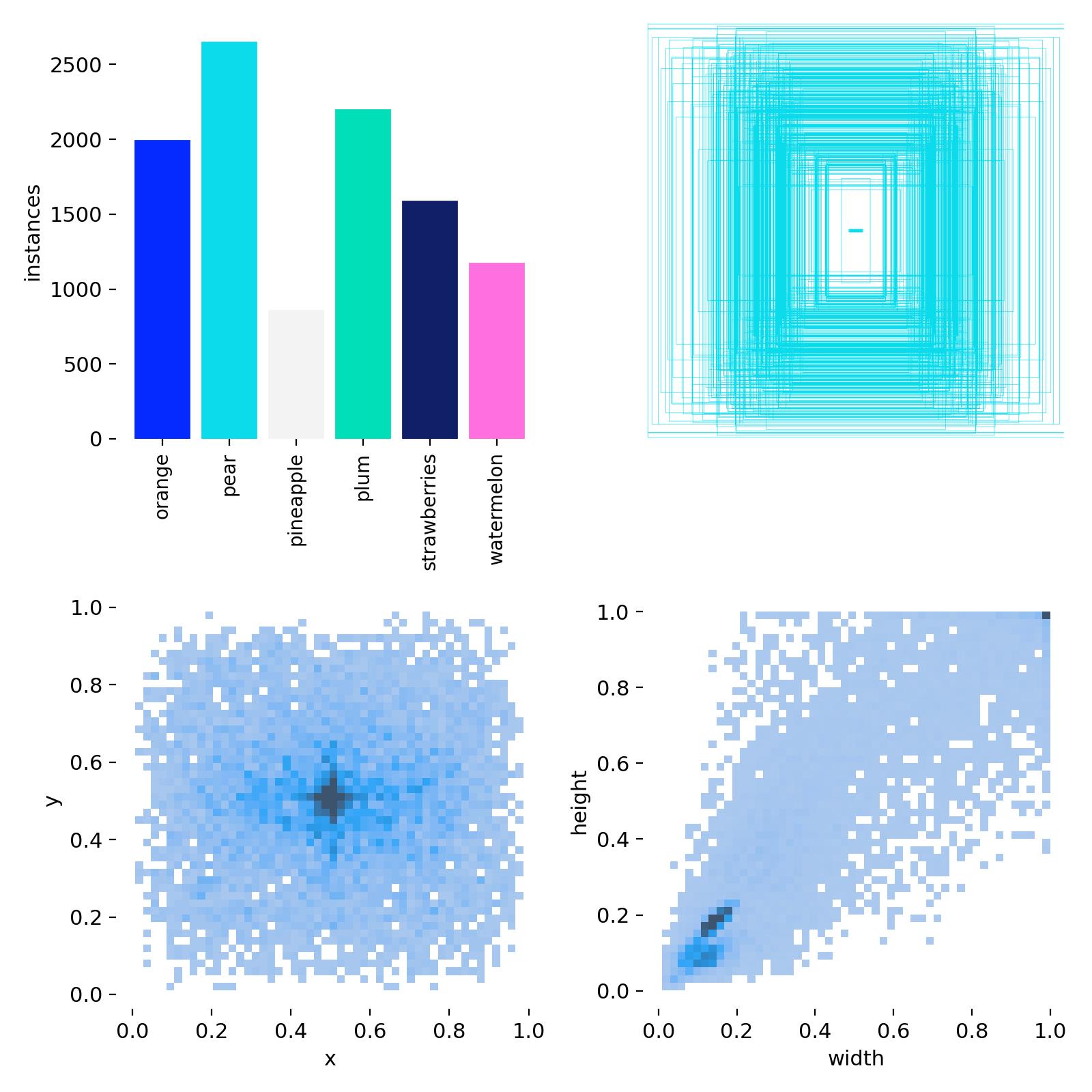}
    \includegraphics[width=0.45\textwidth]{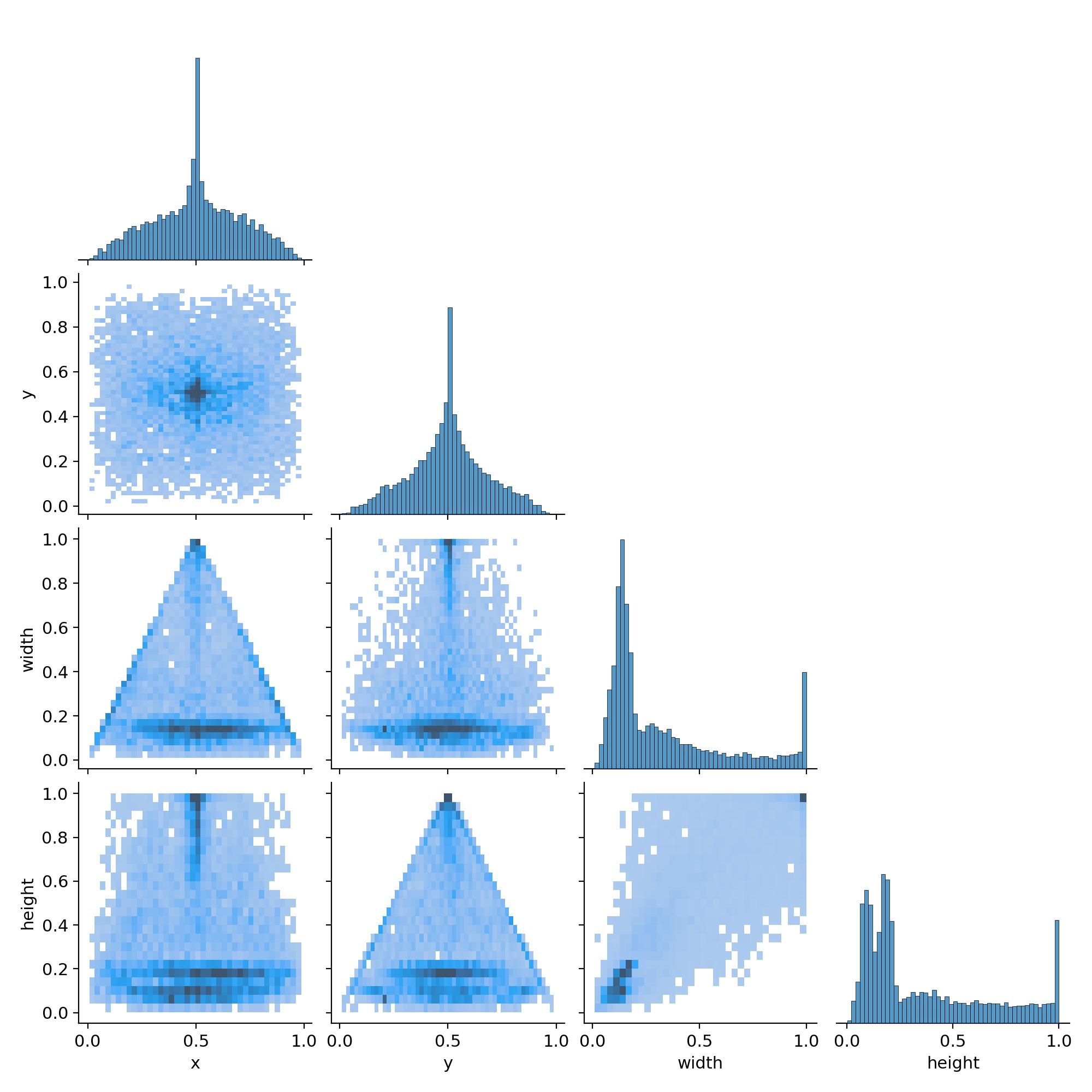}
    \caption{Target Fine-Grained Fruit Dataset Characteristics. (a) Distribution of object instances per class. (b) Correlogram showing bounding box center (x, y), width, and height distributions and relationships.}
    \label{fig:fruit_dataset}
\end{figure}

\subsection{Fine-tuning Protocol}

We investigate the impact of fine-tuning depth by systematically varying the number of trainable layers in the YOLOv8n backbone. Using the \texttt{freeze} parameter within the Ultralytics framework~\cite{Ref-YOLOv8Ultralytics}, we created three experimental conditions:

\begin{itemize}
    \item \textbf{Freeze=22}: Only the final detection head layers (equivalent to layer 22 in the standard architecture) were trained. The entire backbone and neck (layers 0--21) remained frozen with their original COCO pre-trained weights.
    \item \textbf{Freeze=15}: Layers 15 through 21 (later stages of the backbone/neck) and the detection head were fine-tuned. Layers 0--14 remained frozen.
    \item \textbf{Freeze=10}: Layers 10 through 21 and the detection head were fine-tuned. Layers 0--9 (early backbone) remained frozen.
\end{itemize}

During fine-tuning, a callback mechanism ensured that Batch Normalization layers within the frozen segments of the network were kept in evaluation mode to prevent their running statistics from being updated. All models were fine-tuned on the fruit dataset for 100 epochs using an image size of 640$\times$640. Default hyperparameters from the Ultralytics library were used, including the AdamW optimizer~\cite{Ref-ImprovedAdam}. Training was performed on NVIDIA Tesla T4 GPUs.

\subsection{Evaluation Strategy}

We evaluated the models resulting from each fine-tuning protocol on two distinct tasks:

\paragraph{Target Task (Fruit Detection):} Performance on the primary fine-grained task was assessed by evaluating the best model checkpoint (based on validation loss during fine-tuning) from each freeze condition (Freeze=10, 15, 22) on the designated validation split of the fruit dataset. Standard object detection metrics, specifically mean Average Precision at an IoU threshold of 0.5 (mAP@0.5) and the primary COCO metric averaged over IoU thresholds from 0.5 to 0.95 (mAP@0.5:0.95), were calculated.

\paragraph{Source Task (COCO - Catastrophic Forgetting):} To measure the impact of fine-tuning on the model's original capabilities, we evaluated performance on the standard COCO 2017 validation set (val2017). For each freeze condition, we conceptually combined the original COCO detection capability (represented by the unmodified parts of the network and the original head logic) with the newly fine-tuned fruit detection capability into a unified inference model. This combined model was used to generate predictions on val2017. Only predictions corresponding to the 80 original COCO classes were considered. These predictions were formatted according to COCO guidelines and evaluated against the official \texttt{instances\_val2017.json} annotations using the standard \texttt{pycocotools} evaluation suite~\cite{Ref-PyCoTools}. We report the standard COCO AP metrics (AP / mAP@0.5:0.95, AP@0.50, AP@0.75).

A comparison between the backbone weights of the fine-tuned models and the original base model was also performed post-hoc to verify that weight modifications from fine-tuning were successfully saved and loaded during this evaluation.

This dual evaluation allows us to directly assess the trade-off between specializing for the new fine-grained task and retaining performance on the original general-purpose task.

\section{Experiments \& Results}
\label{sec:weight_analysis}

This section details the outcomes of our experiments, evaluating the performance of the YOLOv8n model fine-tuned under different freeze conditions on both the target fine-grained fruit dataset and the source COCO dataset.

\subsection{Experimental Setup and Metrics}

As outlined in Section~3, we conducted experiments using three fine-tuning configurations by setting the \texttt{freeze} parameter to 22, 15, and 10, progressively unfreezing more backbone layers. Performance on the target fruit dataset was measured using mean Average Precision at IoU thresholds of 0.5 (mAP@0.5) and 0.5:0.95 (mAP@0.5:0.95). Performance on the source COCO dataset (val2017) was measured using the standard COCO metrics calculated via \texttt{pycocotools}, primarily focusing on AP / mAP@0.5:0.95.

\subsection{Fruit Detection Performance}

The performance results on the validation set of our custom fine-grained fruit dataset are summarized in Table~\ref{tab:fruit_results}. A clear trend emerges: progressively unfreezing and fine-tuning deeper layers of the YOLOv8n backbone leads to substantial improvements in detection accuracy for this specialized task. The configuration with \texttt{Freeze=10}, which allows adaptation of features from layer 10 onwards, achieves the highest performance, reaching 77.3\% mAP@0.5 and 54.1\% mAP@0.5:0.95. This represents a significant gain of approximately +10\% absolute mAP@0.5 and +10\% absolute mAP@0.5:0.95 compared to the baseline \texttt{Freeze=22} configuration, where only the final head layer was trained.

\begin{figure}[h]
    \centering
    \includegraphics[width=0.95\textwidth]{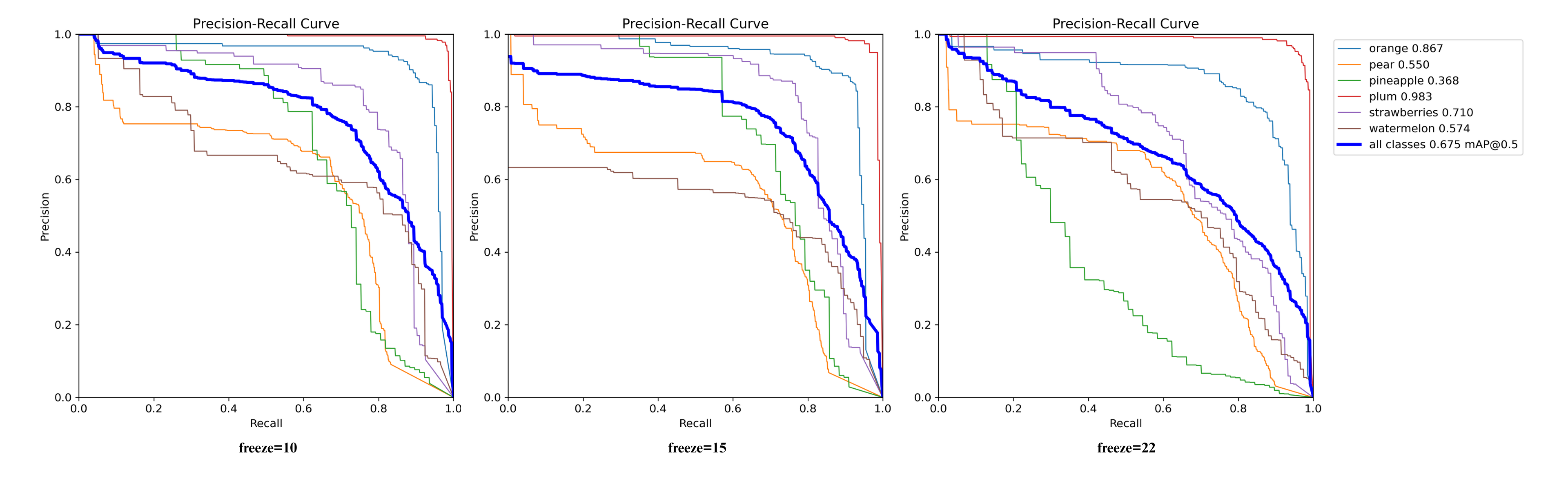}
    \caption{Precision-Recall curves on the fruit validation set (all classes, IoU=0.5) for different fine-tuning freeze levels. The area under the curve (mAP@0.5) increases as more backbone layers are fine-tuned (\texttt{Freeze=10} > \texttt{Freeze=15} > \texttt{Freeze=22}).}
    \label{fig:pr_curves}
\end{figure}

This performance improvement across different confidence thresholds is visually confirmed by the Precision-Recall (PR) curves presented in Figure~\ref{fig:pr_curves}. The curve corresponding to \texttt{Freeze=10} consistently sits above those for \texttt{Freeze=15} and \texttt{Freeze=22}, indicating superior precision at equivalent recall levels (or higher recall at equivalent precision), particularly evident in the higher area under the curve which relates to the mAP scores. This highlights the benefit of adapting mid-to-late backbone features for capturing the fine-grained visual distinctions required by the fruit dataset.

\begin{table}[h]
\centering
\caption{Performance comparison on the fine-grained fruit validation set for different backbone freeze levels. Deeper fine-tuning (lower freeze level) significantly improves detection accuracy. Best results are highlighted in bold.}
\label{tab:fruit_results}
\begin{tabular}{|c|c|c|c|}
\hline
\textbf{Freeze Level} & \textbf{Fine-tuned Layers} & \textbf{Fruit mAP@0.5} & \textbf{Fruit mAP@0.5:0.95} \\
\hline
22 & Head Only (L22)      & 0.675 & 0.443 \\
15 & Layers 15--22        & 0.752 & 0.523 \\
10 & Layers 10--22        & \textbf{0.773} & \textbf{0.541} \\
\hline
\end{tabular}
\end{table}

\subsection{COCO Detection Performance (Forgetting Analysis)}

To assess the impact of fine-tuning on the model's original capabilities, we evaluated the performance of the merged models (incorporating both COCO and fine-tuned fruit heads) on the COCO 2017 validation set. The results are presented in Table~\ref{tab:coco_results}.

The baseline performance, established by the \texttt{Freeze=22} configuration (where the backbone and COCO head remain effectively untouched), is 36.7\% mAP@0.5:0.95 in our experimental setup. Strikingly, the evaluation results show that fine-tuning the backbone deeper (\texttt{Freeze=15} and \texttt{Freeze=10}) resulted in no measurable degradation in performance on the COCO task. All configurations yielded virtually identical mAP scores.

This indicates that, within our experimental setup, significant performance gains on the specialized fine-grained task were achieved without inducing catastrophic forgetting of the model's general object detection capabilities acquired during pre-training.

\begin{table}[h]
\centering
\caption{Performance comparison on the COCO 2017 validation set (val2017) using the merged models from different fruit fine-tuning freeze levels. Performance remains stable, indicating negligible catastrophic forgetting.}
\label{tab:coco_results}
\begin{tabular}{|c|c|c|}
\hline
\textbf{Freeze Level} & \textbf{Fine-tuned Layers} & \textbf{COCO AP / mAP@0.5:0.95} \\
\hline
22 & Head Only (L22)      & 0.367 \\
15 & Layers 15--22        & 0.367 \\
10 & Layers 10--22        & 0.367 \\
\hline
\end{tabular}
\end{table}

\subsection{Weight Verification Summary}

To ensure that the observed lack of forgetting was not an artifact of the fine-tuned weights failing to persist through the model saving and loading process, we performed a direct comparison of the backbone weights (layers 0--21). We compared the weights extracted from the loaded merged model (e.g., \texttt{merged\_freeze\_15.pt}) against the weights from the original base \texttt{yolov8n.pt}.

This comparison revealed numerous significant differences in the weight values for layers within the fine-tuned range (e.g., layers 15--21 for the \texttt{Freeze=15} model), while layers outside this range remained identical. This confirms that the modifications learned during fine-tuning were indeed successfully saved and loaded, validating the COCO performance results reported in Table~\ref{tab:coco_results}.

\section{Discussion and Analysis}
Our experimental results offer valuable insights into the dynamics of transfer learning when adapting a general-purpose object detector like YOLOv8n to a specialized, fine-grained domain. We discuss the implications regarding feature adaptation, catastrophic forgetting, practical fine-tuning strategies, and the limitations of this study.

\subsection{Feature Granularity and Target Task Performance}
The significant improvement observed in fruit detection performance as we unfroze deeper backbone layers (Table~\ref{tab:fruit_results}, Figure~\ref{fig:pr_curves}) strongly supports the hypothesis that effective fine-grained recognition often requires adaptation beyond the final classification head. Pre-trained models learn a hierarchy of features: early layers capture general low-level patterns (edges, textures), mid-layers assemble more complex motifs and parts, and later layers typically encode higher-level semantic information relevant to the pre-training classes~\cite{Ref-ZeilerFergus, Ref-YosinskiFT}.

For the fine-grained fruit dataset, distinguishing between visually similar classes like pears and oranges, or identifying specific textures of pineapples and strawberries, likely relies heavily on adapting the mid-to-late level feature representations (roughly corresponding to layers 10--21 in our setup). Relying solely on the frozen features and adapting only the final head (Freeze=22) proved suboptimal, suggesting the highest-level COCO features were insufficient or required significant repurposing. By allowing fine-tuning of layers 15--21 (Freeze=15) and especially 10--21 (Freeze=10), the model could refine these crucial mid-to-late level features -- adapting representations of shape, texture, and parts -- to better align with the specific discriminative requirements of the fruit classes, leading to the observed substantial gains in mAP.

\subsection{Absence of Catastrophic Forgetting}
Perhaps the most striking result of our study is the negligible catastrophic forgetting observed on the original COCO task (Table~\ref{tab:coco_results}). Despite significant modifications to the backbone weights during fine-tuning (confirmed by our weight verification analysis, Section~\ref{sec:weight_analysis}), particularly in the Freeze=10 and Freeze=15 configurations which yielded the best fruit performance, the models retained their original COCO detection performance (approx.\ 36.7\% mAP@0.5:0.95). This outcome contrasts with the common expectation that adapting a network to a new task, especially through deeper fine-tuning, often leads to a measurable decline in performance on the original source task~\cite{Ref-ArchitectureCL, Ref-McCloskeyForgetting}.

While the exact mechanisms underlying this resilience are complex and beyond the direct scope of this empirical study, we hypothesize several potential contributing factors:

\begin{itemize}
  \item \textbf{Model Capacity and Parameter Redundancy:} Modern architectures like YOLOv8n possess considerable capacity. It is plausible that the network contains sufficient representational power and potentially redundant parameters, allowing subsets of weights to be adapted for the new fruit task without critically disrupting the parameters primarily responsible for COCO object detection.
  \item \textbf{Task Dissimilarity and Feature Specialization:} The visual features crucial for detecting the 80 diverse COCO categories might be sufficiently distinct from the specific, fine-grained features adapted for the 6 fruit classes. The fine-tuning updates might have occurred in parts of the feature or parameter space less critical for the COCO task.
  \item \textbf{Optimization Dynamics:} The fine-tuning process itself (using standard optimizers and learning rates for a fixed number of epochs) might have implicitly constrained the weight updates, preventing drastic shifts away from the original COCO solution manifold while still allowing sufficient adaptation for the fruit task.
  \item \textbf{Architecture and Evaluation:} While our weight comparison confirmed backbone changes persisted, the specific dual-head approach used for evaluation ensures that the original COCO head logic remains separate during inference, though it relies on the potentially modified backbone features.
\end{itemize}

Further investigation would be needed to disentangle these factors definitively. However, the empirical result itself -- achieving significant specialized gains without source task degradation is noteworthy.

\subsection{Practical Implications}
These findings have direct practical implications for applying transfer learning:

\begin{itemize}
  \item \textbf{Reconsidering the Fear of Deeper Fine-tuning:} Practitioners often default to freezing most of the backbone to mitigate catastrophic forgetting and reduce training time. Our results suggest that, at least in some scenarios, the risk of forgetting might be lower than anticipated.
  \item \textbf{Prioritizing Target Performance:} When maximum performance on a specialized, complex, or fine-grained target domain is the priority, exploring deeper fine-tuning strategies (e.g., unfreezing mid-to-late backbone layers) appears warranted and may not necessarily incur a significant penalty on general capabilities. The potential performance gains on the target task, as demonstrated here (+10\% mAP), can be substantial.
\end{itemize}

This suggests a more nuanced approach where the depth of fine-tuning is treated as a key hyperparameter to be explored, especially when adapting models to domains significantly different from or more granular than the original pre-training data.

\subsection{Limitations}
This study, while providing clear results, has limitations that bound the generalizability of its conclusions:

\begin{itemize}
  \item \textbf{Model and Dataset Specificity:} Our findings are based on a single base architecture (YOLOv8n) and one specific fine-grained fruit dataset derived from ~\cite{Ref-FridgeDataset}. The observed balance between adaptation and forgetting might differ with other architectures, pre-training datasets, or target domains.
  \item \textbf{Evaluation Methodology:} The COCO evaluation relied on a specific dual-head inference approach; while weight comparisons confirmed backbone changes, the exact performance interaction might differ slightly with alternative evaluation strategies.
  \item \textbf{Scope of Fine-tuning:} We explored varying the freeze point but did not investigate other fine-tuning aspects like learning rate schedules, regularization techniques, or alternative optimization algorithms, all of which could influence the results.
  \item \textbf{No Explicit Forgetting Mitigation:} We did not employ techniques explicitly designed to prevent catastrophic forgetting (e.g., EWC, LwF). The lack of forgetting was an observed outcome of standard fine-tuning in this context.
\end{itemize}

Therefore, while our results present a compelling case study, further research across diverse models, datasets, and fine-tuning configurations is needed to establish the broader applicability of these findings.

\section{Conclusion}
This study investigated the impact of fine-tuning depth when adapting a pre-trained YOLOv8n object detector to a specialized, fine-grained fruit recognition task. By systematically varying the number of unfrozen backbone layers during fine-tuning, we aimed to understand the trade-offs between performance gains on the target task and the potential for catastrophic forgetting on the original source task (COCO).

Our empirical results provide two key findings. Firstly, we demonstrate that deeper fine-tuning, involving the adaptation of mid-to-late backbone feature extractors (unfreezing down to layer 10), yields substantial improvements in detection accuracy (+10\% absolute mAP@0.5:0.95) on the fine-grained fruit dataset compared to fine-tuning only the final head. Secondly, and perhaps more significantly, we found that this enhanced specialization was achieved with negligible degradation in performance on the original COCO validation set across all tested fine-tuning depths.

This empirical evidence highlights the effectiveness of adapting deeper, more complex features for fine-grained specialization within a standard detector architecture. Furthermore, it suggests that significant adaptation to a new domain is possible without incurring the often-feared penalty of catastrophic forgetting, indicating this trade-off may be more favorable than commonly assumed in certain transfer learning contexts. Future work should explore the generalizability of these findings across a wider range of model architectures, target datasets, and fine-tuning methodologies to further understand the complex interplay between feature adaptation and knowledge retention in deep neural networks.

\end{document}